\documentclass[runningheads]{llncs}

\usepackage{eccv}

\usepackage{eccvabbrv}

\usepackage{graphicx}
\usepackage{booktabs}
\usepackage{soul}

\usepackage[accsupp]{axessibility}  % Improves PDF readability for those with disabilities.
\usepackage{multirow}
\usepackage{makecell}
\usepackage{hyperref}

\usepackage{orcidlink}

\begin{document}

% ---------------------------------------------------------------
\title{Pose-independent 3D Anthropometry from Sparse Data} 

\titlerunning{Pose-independent 3D Anthropometry}

\author{David Bojanić\inst{1,2}\orcidlink{0000-0002-2400-0625} \and
Stefanie Wuhrer\inst{2}\orcidlink{0000-0001-8085-5479} \and
Tomislav Petković\inst{1}\orcidlink{0000-0002-3054-002X} \and \\
Tomislav Pribanić\inst{1}\orcidlink{0000-0002-5415-3630}$^\dagger$}

% $^\cross[.4pt]$

\authorrunning{D.~Bojanić et al.}

\institute{University of Zagreb Faculty of Electrical Engineering and Computing, Zagreb, Croatia \\
% \email{\{first.last\}@fer.unizg.hr}
\email{\{david.bojanic, tomislav.petkovic.jr, tomislav.pribanic\}@fer.unizg.hr}
% Princeton University, Princeton NJ 08544, USA 
\and
Inria centre at the University Grenoble Alpes, Grenoble, France \\
\email{\{first.last\}@inria.fr}}
% Springer Heidelberg, Tiergartenstr.~17, 69121 Heidelberg, Germany

\maketitle
\begin{center}
    %\centering
    \includegraphics[width=0.95\textwidth]{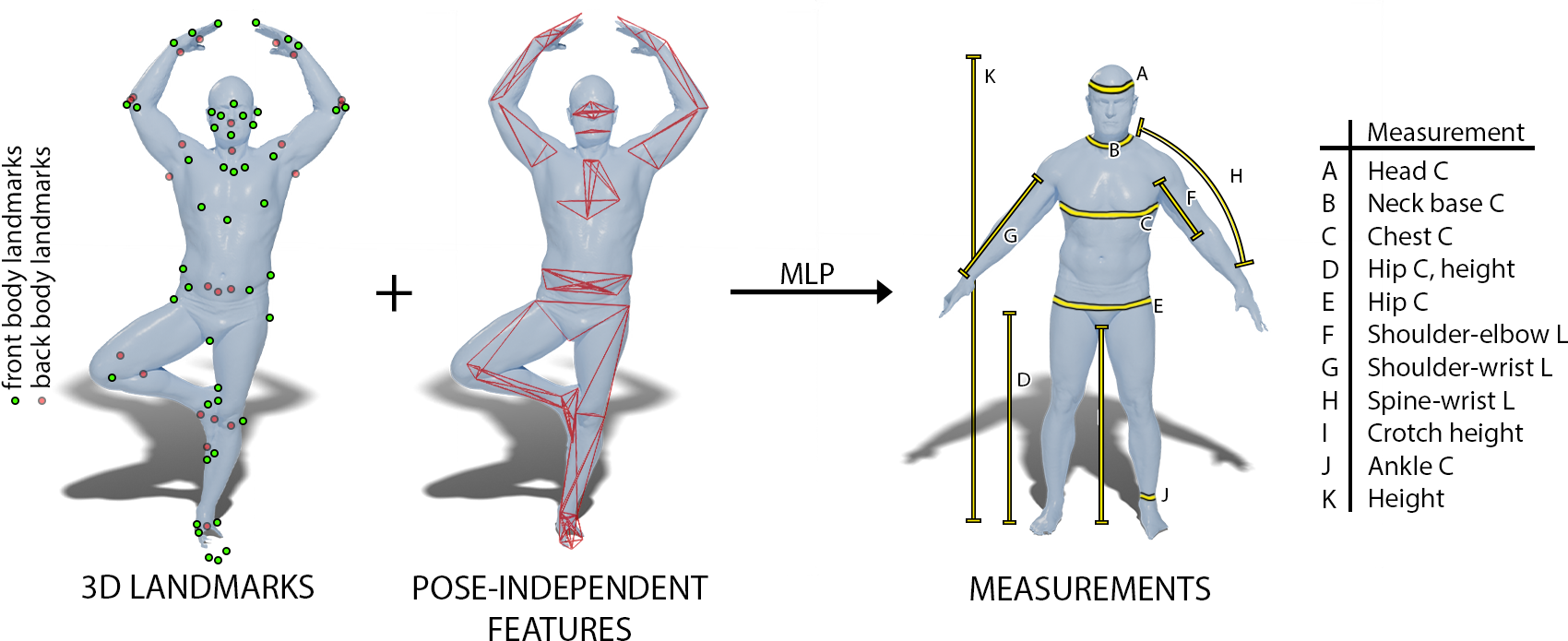}
    \captionof{figure}{We propose a new approach to estimate body measurements from 3D landmark locations on a body in any given pose. We identify pose-independent features that have an impact on the measurements by analyzing the landmarks of a large database of posed scans. The $11$ body measurements listed on the right, are estimated from the $70$ landmarks listed on the left, along with $158$ pose-independent features in the middle.}
    \label{fig:teaser}
\end{center}

\begin{abstract}
  % 1. what is anthropomeatry
  3D digital anthropometry is the study of estimating human body measurements from 3D scans. 
  % 2. why does it matter
  Precise body measurements are important health indicators in the medical industry, and guiding factors in the fashion, ergonomic and entertainment industries.
  % 3. what do others do
  The measuring protocol consists of scanning the whole subject in the static A-pose, which is maintained without breathing or movement during the scanning process. However, the A-pose is not easy to maintain during the whole scanning process, which can last even up to a couple of minutes.
  % A-pose as he remains still and holds his breath during the process. 
  %Then, the obtained scan is processed by removing the scan parts that are not part of the body. 
  This constraint affects the final quality of the scan, which in turn affects the accuracy of the estimated body measurements obtained from methods that rely on dense geometric data. Additionally, this constraint makes it impossible to develop a digital anthropometry method for subjects unable to assume the A-pose, such as those with injuries or disabilities.
  %automate body measurement estimation for some subjects, such as those with disabilities or those prone to fidgeting. 
  % 4. what do we solve?
  We propose a method that can obtain body measurements from sparse landmarks acquired in any pose.
  % 5. how do we solve it?
   We make use of the sparse landmarks of the posed subject to create pose-independent features, and train a network to predict the body measurements as taken from the standard A-pose. 
  % We use a two-step pipeline in which we first detect 3D anatomical landmarks on the body of the subject, which are used in the second step to estimate the body measurements as if they were measured in the standard A-pose.
  % 6. comment on results
  We show that our method achieves comparable results to competing methods that use dense geometry in the standard A-pose, but has the capability of estimating the body measurements from any pose using sparse landmarks only.
  % 7. WE MAKE IT OPEN SOURCE
  Finally, we address the lack of open-source 3D anthropometry methods by making our method available to the research community at %\url{github.com/DavidBoja/PosedLandmarks2Anthropometry}
  \url{github.com/DavidBoja/pose-independent-anthropometry}.
\end{abstract}

\section{Introduction}
\label{sec:introduction}

%1. general anthropometry and why do we want it
Anthropometry is the scientific study of measurements and proportions of the human body \cite{review-body-meas-using-3d-scanning}. It encompasses the whole process from data collection, documentation, summarization, and analysis of body measurements \cite{anthropometry-clothing-book}. These measurements are essential in quantifying the variations in populations and comparing populations of different countries, and ages \cite{caesar-survey, size-uk-survey,development-of-anthropo-work-environment-for-taiwanese-workers}, and it impacts a variety of industries, such as: medicine \cite{digital-anthropometry-critical-review,donlic-phd}, surveying \cite{survey-book,anthropometry-clothing-book}, fashion \cite{anthropometry-clothing-book}, fitness \cite{anthropometric-measurement-in-fitness}, and entertainment \cite{augmented-reality}.

% manual measurements became a pattern for digital ones
Manual anthropometry has been standardized \cite{iso7250-1:2017} in order to achieve comparability between different subjects with different shapes. The procedure is carried out by an anthropometrist with the help of various tools, such as tape measures and calipers \cite{anthropometry-clothing-book}. The standard protocol is comprised of several steps: (1) posing the subject in order to mitigate the influence of the pose onto the final measurement estimation; (2) finding the subject's landmarks which are used as anchors to guide the measurement estimation and; (3) taking the measurement by using the appropriate tool to do so. As an example of taking the chest circumference, the subject is asked to stand upright and exhale, while the tape measure is wrapped around the torso passing through the left and right Thelion landmarks \cite{iso7250-1:2017}. The manual anthropometry standards are the main constraint for efficient and widely available anthropometry since anthropometrists need to be trained in order to estimate accurate measurements \cite{ansur_study,ansur_study2}, and obtaining the measurements takes significant time \cite{review-body-meas-using-3d-scanning}.

%  digital anthropometry follow these standards
With digital anthropometry, the expert anthropometrists are replaced with digital devices, such as 3D scanners, mobile phones and various sensors. The standards, however, remain \cite{iso20685-1:2018}, but are mostly shifted onto the subject being measured instead of the one doing the measuring. These standards are related to the 3D scanning process and are the main constraints of 3D digital anthropometry. In order to scan a subject for measurement estimation, they need to stand in the fixed A-pose (upright standing pose with feet and hands slightly apart) without moving during the scanning, which can last up to 5 minutes with handheld scanning devices \cite{review-body-meas-using-3d-scanning}.

% The standards, however, remain \cite{iso20685-1:2018}. These standards are related to 3D scanning the subject and are the main constraints with 3D digital anthropometry. In order to scan a subject for measurement estimation, he needs to stand in the fixed A-pose (upright standing pose with feet and hands slightly apart) without moving for a short period of time, which can last up to 5 minutes with handheld scanning devices \cite{review-body-meas-using-3d-scanning}.

% The standardization from manual anthropometry has been transferred to digital anthropometry as well \cite{iso20685-1:2018}. However, these standards are also the main constraints for estimating body measurements more easily. The main problem 

% The standard procedure consists of firstly finding the subject's landmarks from which the measurements can be estimated in several ways. One group of approaches finds the measurements as cross-sections of the scan and planes defined using the given landmarks, and a second group of approaches uses the found landmarks to fit a statistical human body model \cite{SMPL,SMPLX} (template) to the given scan, after which the measurements can be transferred from the template body onto the scan. 

% 4. why is that a problem?
This is a great obstacle to having an efficient and more accessible anthropometry method, as even the subject's breathing can affect the precision of the final measurements \cite{review-body-meas-using-3d-scanning,breathing-life-into-shape-capturing-modeling-and-animating-3d-human-breathing}. Since most methods require dense geometric data in order to estimate the measurements, this hinders their performance.  Furthermore, requiring the fixed A-pose in order to estimate the body measurements excludes the subjects with injuries and disabilities \cite{automation-of-full-size-wheelchair-user-body-3d-scan-dim-signs-registration,collection-of-anthropo-from-older-and-physically-impaired-persons-traditional-methods-vs-tc2-3d-body-scanner} or those prone to fidgeting \cite{review-body-meas-using-3d-scanning}.

% 4a mention works that do this -> confirmation that problem is relevant and justify motivation
% WORKS THAT ESTIM POSED ANTHROPO
% 1. Estimating Anthropometry with Microsoft Kinect - motivation is to observe what happens
% 2. A methodology to obtain anthropometric measurements from 4D scans - motivation je dobiti drugacije mjere jer za ergonomiju neki proizvodi mjere recimo "reaches", ili za tight fitting trebaju znati variability of shape in that place --> oni zele drugacije mjere u svakoj pozi
% 3. Tsoli - motivation je korisitti vise scanova covjeka za dobiti bolju procjenu

% 4. efficient model-free .. - motivation is to show they can estimate meas in diff poses, cause their model allows it
% 5. Accurate estimation of body height from a single depth image via a four-stage developing network - idea is to estim height from depth at any pose - height is important, but time consuming to do manually, especially if person is chld or cannot stand up straight or injured

Most of the methods that estimate body measurements of posed subjects, do not generalize well to poses diverging from the standard A-pose \cite{efficient-model-free-anthropometry-from-depth-data,estimating-anthropo-with-microsoft-kinect}, can only generalize to the sitting pose \cite{tsoli-mode-based-anthropometry} or purposefully diverge from the A-pose measurements to model the change in shape of a dynamic sequence \cite{a-methodology-to-obtain-anthropo-meas-from-4d-scans}.

% 5. what do we propose?
To alleviate these constraints, we propose a simple method that can estimate the body measurements from sparse landmarks in any pose. To do so, we identify features that are pose-independent and have an impact on the measurements, by analyzing a large database of posed scans. Using the landmark coordinates and the identified features we estimate the measurements as in the standard A-pose. %With the recent development of 3D scanners, now available even in mobile devices \cite{arkittrack}, this would allow for a widely spread anthropometry method without requiring trained personnel, expensive equipment, and trained subjects.
The 3D body landmark coordinates can be estimated from posed body scans using existing approaches, such as \cite{semi-automatic-prediction-of-landmarks-on-human-models-in-varying-poses,automatic-location-and-semantic-labeling-of-lanmarks-on-3d-human-body-models,automatic-3d-human-body-landmarks-extraction-and-measurement-based-on-mean-curvature-skeleton-for-tailoring}. In this work we assume the landmarks are given and focus on estimating the measurements. This assumption is in line with the current literature, since most of the methods assume landmarks either for template fitting \cite{tsoli-mode-based-anthropometry,pred-anthropo-meas-from-3d-body-scans-methods-and-eval,anthropometrc-clothing-measurements-from-3d-body-scans,efficient-model-based-anthropo-under-clothing-using-low-cost-depth-sensors,precise-automatic-anthropo-measure-extr-using-template-registration,a-methodology-to-obtain-anthropo-meas-from-4d-scans}, measurement estimation \cite{3d-anthropometric-algo-for-estim-meas-required-for-specialized-garment-design,automatic-3d-human-body-landmarks-extraction-and-measurement-based-on-mean-curvature-skeleton-for-tailoring,automated-anthropometric-data-collection-from-3d-digital-human-models,automated-anthropo-data-collection-using-3d-whole-body-scanners,automatic-body-measurement-based-on-slicing-loops,body2vec,estimating-anthropo-with-microsoft-kinect,anthropo-body-meas-based-on-multi-view-stereo-image-reconstruction,fast-and-automatic-body-circular-meas-based-on-a-single-kinect,rgb-d-cam-based-anthropo-meas-system-for-barong-tagalog-tailoring,development-of-automatic-3d-body-scan-meas-line-generation-method}, or use other similar priors \cite{home-3d-body-scans-from-noisy-image-and-range-data,a-template-based-anthropo-measurement-approach-for-3d-scanned-bodies}.

% 6. criticize open-source
A great challenge with the 3D anthropometry literature are the limited number of open-source datasets and virtually no open-source implementations. Most works create a private small-scale dataset \cite{automatic-body-measurement-based-on-slicing-loops,dl-based-automated-extr-of-antrhopo-meas-from-single-3d-scan,automated-anthropo-data-collection-using-3d-whole-body-scanners,ANET,3d-anthropometric-algo-for-estim-meas-required-for-specialized-garment-design,precise-automatic-anthropo-measure-extr-using-template-registration,pred-anthropo-meas-from-3d-body-scans-methods-and-eval,automatic-3d-human-body-landmarks-extraction-and-measurement-based-on-mean-curvature-skeleton-for-tailoring,automated-anthropometric-data-collection-from-3d-digital-human-models,body2vec,efficient-model-free-anthropometry-from-depth-data,a-template-based-method-for-automatic-anthropo-meas-from-multiple-3d-scans,home-3d-body-scans-from-noisy-image-and-range-data} in order to test their method on several human body scans, and do not share it with the community due to privacy issues. Furthermore, to the best of our knowledge, none of the works, except \cite{bojanic_VISAPP24}, share their code implementations with the community, making the comparison between different methods impossible. 
% We compare our method with  \cite{tsoli-mode-based-anthropometry,anthroscan,scan-db} which share their results on the proprietary CAESAR dataset \cite{caesar-survey} available for purchase. However, we resort to comparing on a different sample of the dataset since no information is provided regarding the train-test subject split.
We compare our method with  \cite{tsoli-mode-based-anthropometry,anthroscan,scan-db,bojanic_VISAPP24} which share their results on the commercial CAESAR \cite{caesar-survey} dataset, and with \cite{efficient-model-free-anthropometry-from-depth-data} on the DYNA \cite{DYNA} dataset.
%, and with \cite{efficient-model-free-anthropometry-from-depth-data} which share their results on the freely available DYNA \cite{DYNA} dataset. 
To make a step towards a more open source 3D anthropometry research, we share the exact subjects and scans we evaluate our method on, and make our implementation available at \url{github.com/DavidBoja/pose-independent-anthropometry}.

\noindent In summary, our contributions are:
\begin{itemize}
    \item A method to estimate body measurements in any given pose using only sparse landmark data.
    \item Making the method and detailed evaluation protocol available to the research community.
\end{itemize}

\section{Related work}
\label{sec:related-work}

We first give a general overview of related works that estimate body measurements, after which we focus on methods that estimate the measurements from posed subjects.

\subsection{3D anthropometry} 

Existing approaches for automatic body measurement estimation from 3D scans fall into three categories: landmark-based, template-based, and direct methods. 

% LANDMARK-BASED METHODS
Landmark-based methods use 3D landmarks and dense scan data to estimate the measurements~\cite{3d-anthropometric-algo-for-estim-meas-required-for-specialized-garment-design,automatic-3d-human-body-landmarks-extraction-and-measurement-based-on-mean-curvature-skeleton-for-tailoring,automated-anthropometric-data-collection-from-3d-digital-human-models,automated-anthropo-data-collection-using-3d-whole-body-scanners,automatic-body-measurement-based-on-slicing-loops,body2vec,estimating-anthropo-with-microsoft-kinect,anthropo-body-meas-based-on-multi-view-stereo-image-reconstruction,fast-and-automatic-body-circular-meas-based-on-a-single-kinect,rgb-d-cam-based-anthropo-meas-system-for-barong-tagalog-tailoring,development-of-automatic-3d-body-scan-meas-line-generation-method}. Length measurements can be estimated as Euclidean or geodesic distances between the landmarks, and circumference measurements can be estimated by cutting the dense scan with a plane at a desired landmark location and finding the cross-section. Rather than cutting the scan with a plane, a path can also be found~\cite{automatic-human-body-feature-extr-and-personal-size-meas} or a curve can be fitted~\cite{rgb-d-cam-based-anthropo-meas-system-for-barong-tagalog-tailoring,body2vec,automated-anthropo-data-collection-using-3d-whole-body-scanners} to a set of landmarks from the body mesh. These methods, however, usually require prior information, such as the subject's sex~\cite{automated-anthropo-data-collection-using-3d-whole-body-scanners}, orientation~\cite{3d-anthropometric-algo-for-estim-meas-required-for-specialized-garment-design,automated-anthropometric-data-collection-from-3d-digital-human-models,automated-anthropo-data-collection-using-3d-whole-body-scanners,anthropo-body-meas-based-on-multi-view-stereo-image-reconstruction}, height~\cite{body2vec}, or pose~\cite{automatic-body-measurement-based-on-slicing-loops,fast-and-automatic-body-circular-meas-based-on-a-single-kinect,estimating-anthropo-with-microsoft-kinect}. Furthermore, these methods require dense scan data since they estimate the measurements directly from the given scan, which makes them sensitive to noise and missing data~\cite{ANET}.

Our method, on the other hand, does not require any prior information, and estimates the measurements using only the landmark locations, therefore avoiding the necessity for dense scan data. Furthermore, by learning the measurements, we can address the method's sensitivity to noise in the data.

% TEMPLATE-BASED METHODS
% 1. template-based - ML / DL - measurements direct
% 2. template-based - fit (traditional or ML / DL) - measurement transfer
% 4. template-based - landmarks - cutting
Template-based methods fit a template to a 3D scan from which the measurement can be learned, transferred, or estimated from 3D landmarks. To learn the measurements~\cite{tsoli-mode-based-anthropometry,pred-anthropo-meas-from-3d-body-scans-methods-and-eval,anthropometrc-clothing-measurements-from-3d-body-scans,scan-db,home-3d-body-scans-from-noisy-image-and-range-data,efficient-model-based-anthropo-under-clothing-using-low-cost-depth-sensors} extract a combination of features from the fitted template (such as the 3D points, predefined paths, shape parameters, mesh edge lengths, or PCA coefficients of triangle deformations) to train different models, such as as the ElasticNet~\cite{elastic-net}, SVR~\cite{SVR} and PLS~\cite{PLS}. To transfer the measurements~\cite{dl-based-automated-extr-of-antrhopo-meas-from-single-3d-scan,ANET,precise-automatic-anthropo-measure-extr-using-template-registration,a-template-based-anthropo-measurement-approach-for-3d-scanned-bodies,a-methodology-to-obtain-anthropo-meas-from-4d-scans} predefine the body measurement paths on the template, which can (optionally) be transferred onto the scan by finding the nearest neighbor of each path point from the template to the scan. Similarly, to estimate the measurements from landmarks,~\cite{im2fit,CALVIS,a-template-based-method-for-automatic-anthropo-meas-from-multiple-3d-scans,anthropo-param-meas-from-equivariant-multi-view-images} transfer the landmarks from the template to the scan, and find the measurement using the landmark-based methods described above. Template-based methods tend to be sensitive to the fitting process~\cite{building-statistical-shape-spaces-for-3d-human-modeling}, and are usually stabilized with additional information such as landmarks~\cite{anthropometrc-clothing-measurements-from-3d-body-scans}, texture maps~\cite{DYNA,DFAUST} or rendered silhouettes~\cite{home-3d-body-scans-from-noisy-image-and-range-data}. Additionally, similarly to landmark-based methods, template-based methods need dense scan data to accurately represent the shape of the subject.

In contrast, our method avoids the fitting process and does not need dense scan data to estimate the measurements.

% DIRECT METHODS
Direct methods learn body measurements from the 3D scans. Whereas~\cite{automatic-estim-anthropo-human-body-meas,inception-network-for-anthropo-body-meas-estim-from-structured-point-cloud,anthroscan} learn to predict the measurements from the 3D points,~\cite{efficient-model-free-anthropometry-from-depth-data} uses gradient-boosted trees to predict local measurements, which are weighted in order to compute the final measurements. A more recent method~\cite{bojanic_VISAPP24}, uses the landmark locations to learn a Bayesian regression model that predicts the measurements. These methods resort to training their methods using synthetic datasets created with statistical human body models~\cite{SMPL,building-statistical-shape-spaces-for-3d-human-modeling}. Because of this, they tend to have difficulties in generalizing to real data. 

Compared to direct methods, we train our method by posing 3D human scans with manual ground truth measurement annotations. Additionally, we only use the landmark coordinates as inputs to our method instead of the full scans. This way, we minimize the synthetic-to-real domain gap by utilizing a simpler representation, which can be bridged more easily during training \cite{akash-synthetic-training-for-accurate-3d-human-pose-and-shape-estim-in-the-wild}.

%% ANTHROPOMETRY FROM POSED SUBJECTS
\subsection{3D anthropometry from posed subjects}

Some works estimate anthropometric measurements from poses that differ substantially from the standardized A-pose. 
%The main reason for this is to minimize the discrepancy from the manual measurements, generally adopted as ground truth when formulating a new digital anthropometry method. 
Among landmark-based methods, Robinson and Parkinson \cite{estimating-anthropo-with-microsoft-kinect} estimate only a few length measurements from joint locations predicted from a Microsoft Kinect \cite{real-time-human-pose-recognition-in-parts-from-single-depth-images}. Among template-based methods, Uriel et al. \cite{a-methodology-to-obtain-anthropo-meas-from-4d-scans} extract the measurements directly from the fitted SMPL \cite{SMPL} body model to a sequence of motion, whereas Tsoli et al. \cite{tsoli-mode-based-anthropometry} learn an ElasticNet \cite{elastic-net} model from a set of features extracted from the body model in two poses. Among direct methods, Probst et al. \cite{efficient-model-free-anthropometry-from-depth-data} use gradient-boosted trees to find the measurements from a depth map of the subject, whereas Yin and Zhou \cite{accurate-estim-of-body-height-from-single-depth-image-via-a-four-stage-developing-network} use a CNN to learn the height of the subject. 

Compared to these, our method can estimate multiple body measurements, avoids the template fitting process, does not rely on dense scan data, and generalizes well to different datasets. We compare our approach with approaches from different categories \cite{efficient-model-free-anthropometry-from-depth-data,anthroscan,bojanic_VISAPP24,tsoli-mode-based-anthropometry,scan-db} that provide their results on the CAESAR and DYNA datasets.

\begin{figure}[tb]
  \centering
  \includegraphics[trim={0.5cm 0.2cm 0.1cm 0.2cm},clip,width=0.6\linewidth]{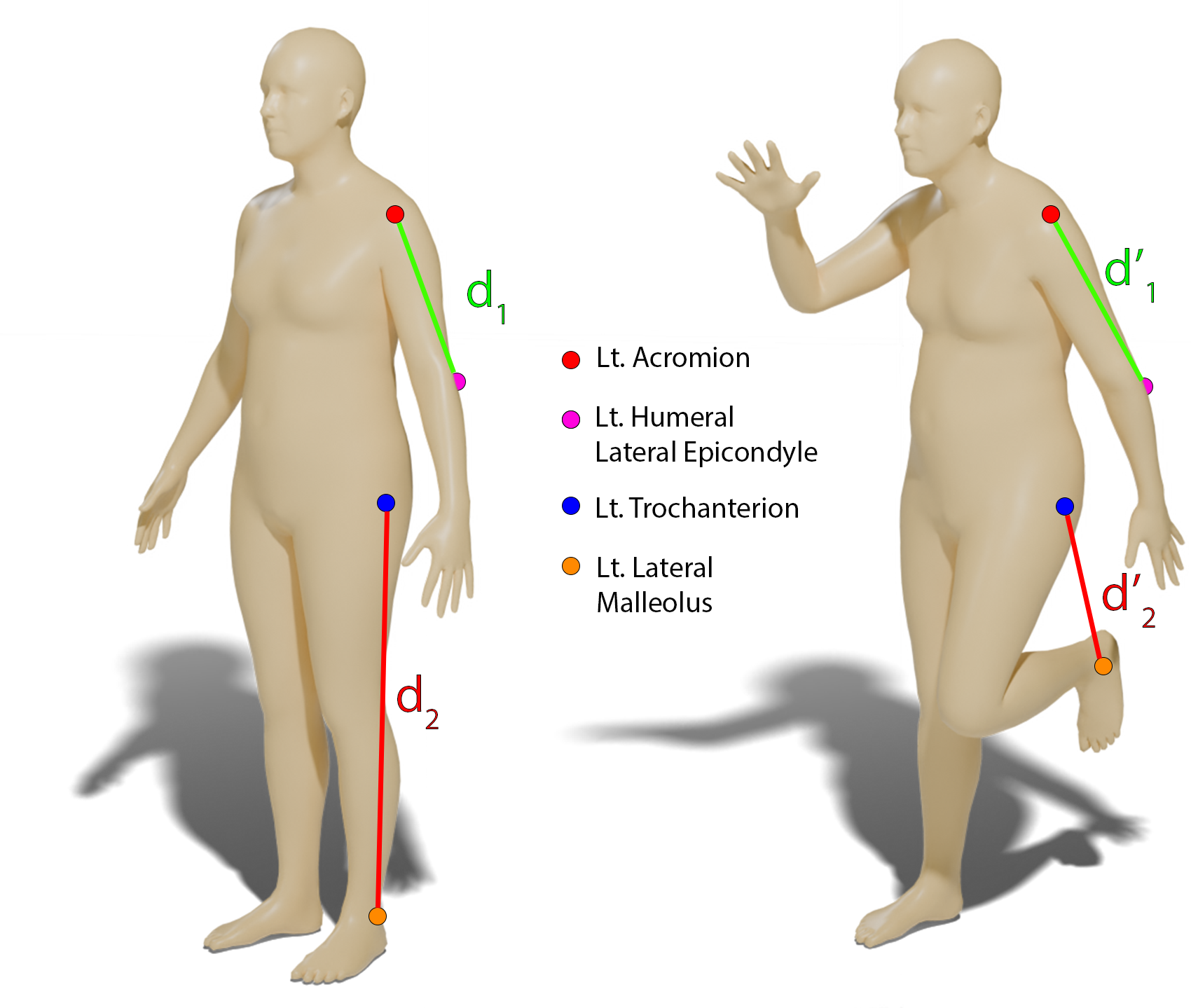}
  \caption{We show the landmark distances of the same subject in two different poses. As can be seen, the same landmark distances $d_1$ and $d'_1$ do not significantly change between the two poses, observing an absolute difference of only $0.78$ cm. On the contrary, the landmark distances $d_2$ and $d'_2$ change significantly because of the articulated deformation of the leg. In this case, the absolute difference between $d_2$ and $d'_2$ is $36.75$ cm. In our method we only use landmark distances where the median difference from its appropriate A-pose distance is less than $1$ cm, such as the distance $d_1$.}
  \label{fig:landmark-distances}
\end{figure}

\section{Method}

We learn a Multilayer Perceptron (MLP) to predict the $11$ body measurements visualized in Fig.~\ref{fig:teaser} (right). The inputs to our method are the $70$ landmarks visualized in Fig.~\ref{fig:teaser} (left). We normalize the landmark coordinates using the pelvis landmarks as follows: in the first step, we find the middle point between the Lt. and Rt. Psis landmarks, which we denominate as Middle Psis. Using this point, along with the Lt. and Rt. Asis landmarks, we create a triangle. The center of the triangle is then translated to the origin, whereas the triangle normal is aligned with the $y$-axis. The normal orientation is pointing towards the half plane containing the Nuchale landmark. Finally, we rotate the landmarks around the $y$-axis to align the Rt. Asis landmark with the $z$-axis. We unravel the normalized coordinates into a $70 \times 3 = 210$ dimensional vector to input into the model.

% \begin{figure}[tb]
%   \centering
%   \includegraphics[width=\linewidth]{figures/landmarks-used.png}
%   \caption{The $70$ body landmarks used to predict the measurements. Landmarks in green are on the front of the body, whilst landmarks in red are on the back of the body.}
%   \label{fig:landmarks-used}
% \end{figure}

To facilitate learning, we additionally identify pose-independent features and provide them to the model. Intuitively, the distances between landmarks located on the same bone of the body should remain constant with pose variation. We show an example of the landmark features in Fig. \ref{fig:landmark-distances}, where the distance between the \textit{Lt. Acromion} (shoulder landmark) and \textit{Lt. Humeral Lateral Epicondyle} (elbow landmark) does not change with pose variation. Contrarily, the distance between the \textit{Lt. Trochanterion} (hip landmark) and the \textit{Lt. Lateral Malleolus} (foot landmark) changes a lot with pose variation since the body deforms in a non-rigid manner. To find the pose-independent landmark distances (features), we pose a single subject in around $410\, 000$ poses, and analyze the differences of the corresponding landmark distances from their reference distance in the A-pose. Details about the posing are provided in Sec. \ref{sec:datasets-CAESAR}.
Finally, we choose the landmark distances that have a median difference less than $1$ cm, which results in $158$ landmark distances. Interestingly, as can be seen from the full set of pose-independent features in Fig. \ref{fig:teaser}, the chosen features are not necessarily symmetric w.r.t. the body, since some of the features from the left (or right) side of the body have been selected, but their respective right (or left) counterparts have not. This happens for the landmark distances that are very close to the cutoff threshold of $1$ cm, where one of the distances is slightly below the threshold, and its counterpart is slightly above. The difference between these counterparts is around $2$ mm.
%where the left (or right) landmark distance is slightly below the threshold, whereas its right (or left) counterpart is slightly above.

The final inputs to our model are comprised of $210$ landmark coordinates and $158$ landmark distances, resulting in a $368$ dimensional input vector. 
We train our model using a MLP network with three layers of dimensions $194$, $97$ and $11$ with a ReLu activation after the first two layers. We optimize the Mean Squared Error (MSE) between the predicted and ground truth $11$ measurements.

\section{Datasets}
\label{sec:datasets}

%We use parts of CAESAR \cite{caesar-survey}, DYNA \cite{DYNA} and parts of the 4DHumanOutfit \cite{4DHumanOutfit} datasets to train and evaluate our method. 
We train and evaluate our method on parts of the CAESAR \cite{caesar-survey} dataset, and evaluate it on parts of the DYNA \cite{DYNA} and 4DHumanOutfit \cite{4DHumanOutfit} datasets. We specify the details of each dataset in the following.

\subsection{CAESAR} \label{sec:datasets-CAESAR}
The CAESAR dataset contains $4396$ subjects scanned in the standing A-pose and two sitting poses. $73$ body landmarks were manually annotated, from which we use the subset of $70$ landmarks visualized in Fig.~\ref{fig:teaser} (left). The subjects were manually measured, and we use a subset of $11$ measurements shown in Fig.~\ref{fig:teaser} (right) following common practice~\cite{tsoli-mode-based-anthropometry}. To compare with existing methods, we use the training / test split by Tsoli~\etal~\cite{tsoli-mode-based-anthropometry} containing $1644$ subjects ($849$ male, $795$ female) for training and $400$ subjects ($200$ male, $200$ female) for testing. Of the training and testing subjects, $120$ and $30$ have at least one missing measurement or landmark, respectively. Therefore, we use the remaining $1424 + 100$ subjects to train and validate, and the remaining $370$ subjects to test our method. Depending on the pose, we train and evaluate on several different subsets of CAESAR, which we explain in the following.

\subsubsection{Posed training set.} To train our method on arbitrary poses, we augment the CAESAR dataset by reposing the $1424$ training subjects into $12000$ different poses. We start by fitting the SMPL model to the A-pose scans. We use a 2-step optimization technique \cite{SMPL-Fitting}. In the first step, we optimize for the pose, shape, scale and translation parameters of the body model, which we supervise using the provided body landmarks. In the second step, we optimize for the SMPL template vertex locations. After fitting, the mean directional chamfer distance from the fitted template to the scans is $0.31$cm, whereas the mean directional chamfer distance from the scans to the fitted templates is $1$cm. We then use the fitted SMPL to unpose and repose the scans into a set of $12000$ poses, comprised of $1000$ standing poses, $1000$ sitting poses and $10000$ varying SMPL poses obtained from several datasets~\cite{amass,up3d,3dpw,h36m}. For each pose, we randomly choose a subject from the CAESAR training subjects, and repose them. Therefore, we train our method using the reposed landmarks and supervise it using the provided manual measurements.

\subsubsection{Standing test set.} We test our method on the standard A-pose provided by the CAESAR dataset. We evaluate on the $370$ testing subjects and use the provided landmark coordinates and manual measurements.

\subsubsection{Noisy test set.} To observe the robustness of our method to landmarking errors, we augment the standing test set described above by moving the landmarks along the mesh up to $0.56$cm, which corresponds to the average median manual inter-observer landmarking error for all the landmarks \cite{errors-in-landmarking-and-the-eval-of-the-acc-of-traditional-and-3d-anthropo}. 
%\hl{Try experiment where each LM is augmented with their own landmarking error} \hl{provide augmentation averages for each LM} 
Each landmark is augmented by moving it along the cross-section of the scan and random plane obtained at the landmark point.

\subsubsection{Sitting test set.} The CAESAR dataset provides scans and landmarks for $4345$ out of the $4396$ subjects in the sitting B-pose. From these, no subject has all of the landmarks necessary to test our method. Therefore, we transfer the missing landmarks from the fitted SMPL template. We evaluate on the same $370$ testing subjects and use the provided manual measurements from the CAESAR dataset.

\subsubsection{Posed test set.} 
\begin{figure}[t]
  \centering
  \includegraphics[trim={0 0 0.1cm 0.5cm},clip,width=\textwidth]{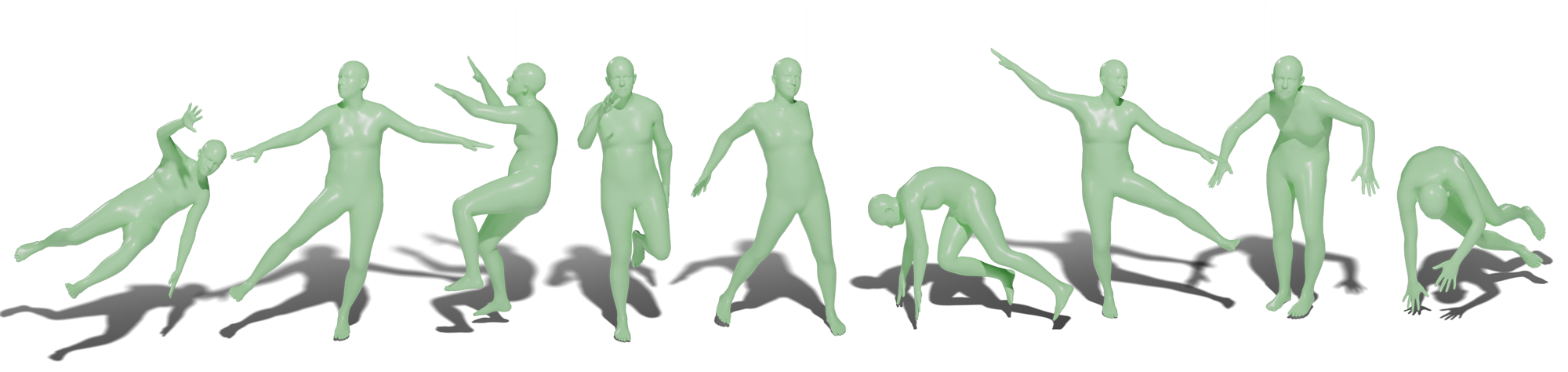}
  \caption{Some of the poses used to create the posed test set. We augment the CAESAR dataset by reposing the subjects in various poses obtained from the AIST \cite{aist_dataset} dataset.}
  \label{fig:caesar-posed-poses}
\end{figure}

To test the pose-independency of our method, we use the same approach as for the training dataset to pose the testing subjects in various poses. We use a sample of $370$ poses from the AIST \cite{aist_dataset} dataset, which is comprised of highly diverse poses obtained from real dancing subjects. Note that these poses have not been seen by the model during training. We visualize some of the poses in Fig.~\ref{fig:caesar-posed-poses}.

\subsection{Dynamic test set} To test the robustness of our method w.r.t. motion, we follow \cite{efficient-model-free-anthropometry-from-depth-data} and estimate the measurements of a single subject during a jumping-jacks sequence from the DYNA \cite{DYNA} dataset (with id $50009$). To obtain the landmarks, DYNA provides accurate SMPL fittings, from which we extract the $70$ body landmarks. Since the ground truth measurements are not provided, we follow \cite{efficient-model-free-anthropometry-from-depth-data} and evaluate the standard deviation of the measurement difference between each frame and the first frame.

\subsection{Clothed test set} To test the robustness of our method w.r.t. clothed subjects, we use a sample of the 4DHumanOutfit \cite{4DHumanOutfit} dataset originally comprised of $18$ subjects performing $10$ actions in $6$ different outfits. We use a subset of $6$ subjects (namely Ben, Leo, Mat, Kim, Mia, and Sue) performing $3$ actions (namely dance, run, avoid) in tight clothing, amounting to $5640$ frames. Similarly to the DYNA evaluation, for each sequence we compute the difference between the first frame measurement estimation and the remaining ones. Then, we compute the standard deviation over these measurement differences for each measurement separately.

\section{Experiments}

% direct
% - Anthroscan - verts -> measurements
% - VISAP work - LM -> measurements in A-pose
% - efficient-model-free-anthropometry-from-depth-data - forest from depth map -> meas

% template-based
% - tsoli - fit smpl -> extract features -> measurements
% - scan-db - fit scape -> verts -> measurements
% - metric-regression-forests-for-human-pose-estim - fit MPII HumanShape model -> measuremnets from mesh

We compare our method with three template-based methods \cite{tsoli-mode-based-anthropometry,scan-db,metric-regression-forests-for-human-pose-estim} and three direct methods \cite{anthroscan,bojanic_VISAPP24,efficient-model-free-anthropometry-from-depth-data}. Among the template-based methods, \cite{metric-regression-forests-for-human-pose-estim,scan-db} estimate the measurements from the fitted template vertices, whereas Tsoli et al. \cite{tsoli-mode-based-anthropometry} extract features from the fitted template in order to estimate the measurements. Among the direct methods, the commercial solution Anthroscan \cite{anthroscan} estimates the measurements directly from the 3D points whereas Probst et al. \cite{efficient-model-free-anthropometry-from-depth-data} estimate them from a depth map of the subject using gradient-boosted trees. 

All of the previously mentioned methods use dense scan data in order to estimate the measurements, and we report their results based on the training and test sets provided by~\cite{tsoli-mode-based-anthropometry}. For the methods operating on sparse landmark data, all tests are effected on the $370$ scans for which all measurements and landmarks in the A-pose are available. The work from Bojani\'{c} et al. \cite{bojanic_VISAPP24} is the only method that uses sparse landmarks in the A-pose to estimate the measurements. However, their approach is not adept to estimate the measurements in any other pose. To mitigate this and have a method to compare with in the \textit{sparse data} category, we introduce a template-based baseline that can estimate the measurements from sparse data in any given pose. The baseline fits an SMPL body model to the given posed landmark coordinates. Next the SMPL is reposed to the A-pose by retaining the fitted shape, and measurements are extracted using the standard landmark-based approach, where distances between landmarks, as well as circumferences of cross-sections are measured, as described above. The baseline provides an insight into the difficulty of the problem, which stems from the ambiguity of using sparse landmark locations to estimate the measurements. 

\begin{figure}[t]
  \centering
  \includegraphics[trim={0.5cm 1cm 1.5cm 2.5cm},clip,width=0.85\textwidth]{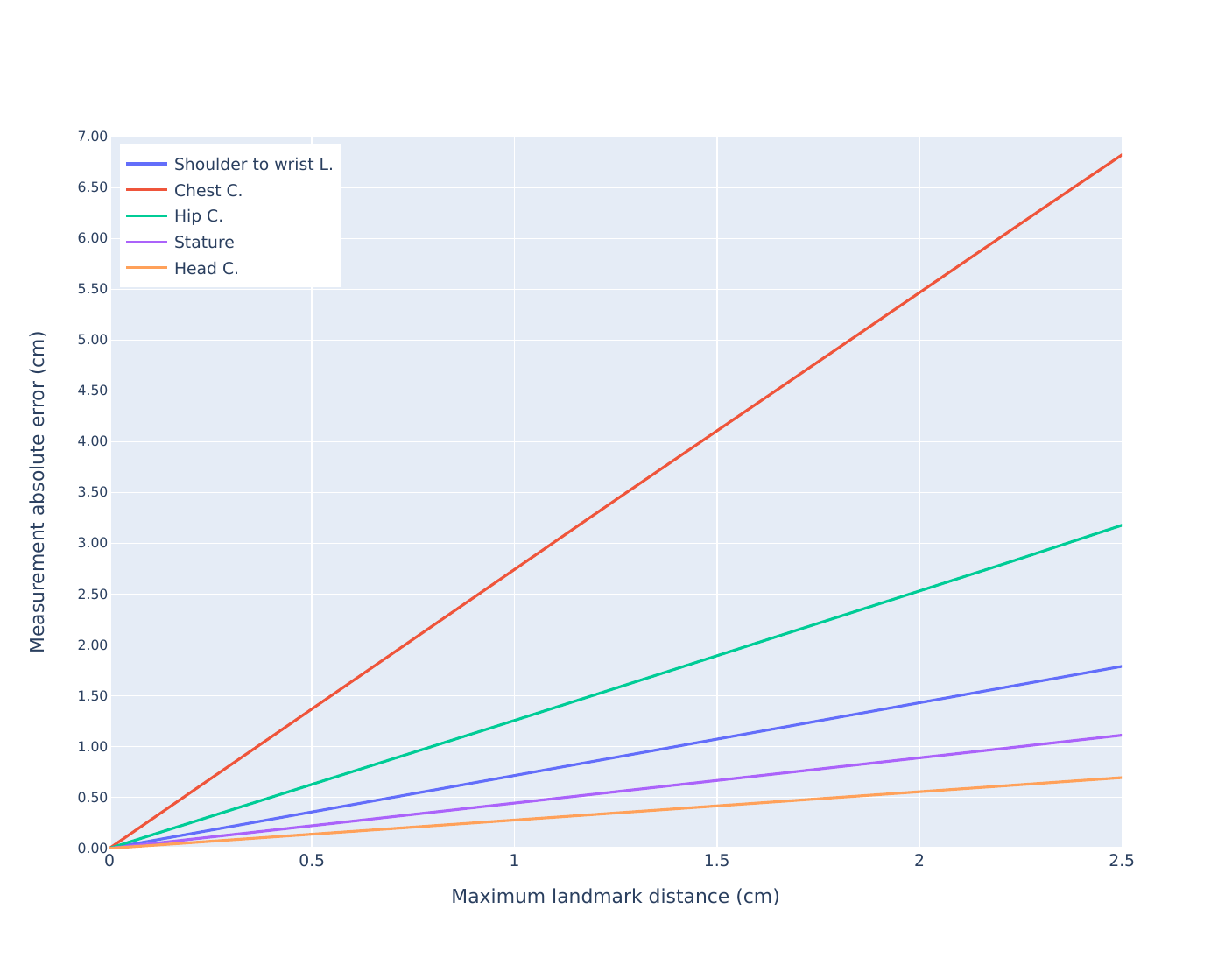}
  \caption{We answer the question whether different body shapes, with different measurements, can share the same landmarks. %As can be seen, the relationship between the landmarks and measurement is linear.
  The relationship between the landmarks and measurements is linear since it is modeled with SMPL, and shows an approximation of the relationship that would be obtained with real data.
  As can be seen, the chest and hip circumference measurements change faster than the landmark distances, indicating that two subjects can have close-by landmark coordinates with different chest and hip circumferences.}
  \label{fig:ambiguity-lm-measurements}
\end{figure}

To further explore this ambiguity, we aim to answer the question whether different body shapes, with different measurements, can share the same landmarks. We use the SMPL model in the fixed neutral pose $\theta_0$ to model the relationship between the landmarks and measurements. We start from a reference shape $\beta_\text{ref}$ obtained by fitting the SMPL model to an arbitrary real subject from the CAESAR dataset. We manually mark the $70$ SMPL landmarks as $L(\theta_0,\beta_\text{ref})$, whose locations depend on the same SMPL pose and shape parameters $\theta$ and $\beta$. 
%Landmarks can be easily extracted from the reference shape by taking the appropriate landmark indices as
% \begin{equation}
%     L(\theta_0,\beta_\text{ref}) = S * \text{SMPL}(\theta_0,\beta_\text{ref})
% \end{equation} 
% where $S \in \{0,1\}^{K \times 6890}$ samples the landmarks from the SMPL vertices.
We optimize for a displaced shape $\beta_\text{ref} + \delta$, that is at unit distant from the reference shape and has the closest landmarks in terms of the Euclidean distance to the reference landmarks. Then, by moving in the direction of the \textit{optimized shape} $\beta_\text{ref} + k \delta$ for various steps $k$, we can observe the variation of the body measurements with almost no landmark variation. More precisely, we optimize:

\footnotesize
\begin{equation}
    \min_\delta \Bigg( \sum_i \quad \bigl\lVert \;\underbrace{L_i(\theta_0,\beta_\text{ref})}_{\substack{\text{\hbox to 1cm{\hss reference landmarks\hss}}}}\quad - \quad\underbrace{L_i(\theta_0,\beta_\text{ref} + \delta)}_{\substack{\text{\hbox to 1cm{\hss optimized landmarks\hss}}}}\; \bigr\rVert_2
    \quad + \quad \underbrace{\bigl\lvert \lVert \delta \rVert_2 - 1 \bigr\rvert}_{\substack{\text{unit norm}}} \Bigg),
\end{equation}
\normalsize
where $L_i$ indicates the $i$-th landmark. Fig.~\ref{fig:ambiguity-lm-measurements} visualizes  for each displacement $k$ the maximum landmark distance between the reference and optimized shapes on the $x$-axis and the corresponding measurement error on the $y$-axis. 

As can be seen, the relative change in body measurements is not proportional to the relative change in landmark displacement. When the maximum landmark distance between the reference and optimized shape reaches $1$cm, the chest circumference, for example, reaches a high $2.7$cm of change. This shows that estimating the body measurements from sparse landmark data is a difficult task since the measurements can be ambiguous for very similar landmark locations. %Furthermore, as we can see from Fig. \ref{fig:lower-bound-analysis}, the shape of the subject, in terms of the median per-vertex-error, changes more drastically than his landmarks, which indicates that the shape information of a subject is not completely contained in their landmarks. Our method tackles these issues by relying on the correlation between different landmark distances and body proportions in order to beat the lower bound baseline. 
Furthermore, the $x$-axis shows the maximum landmark distance up to $2.5$cm, which is an average displacement of vertices that can be caused just by the subject's breathing \cite{breathing-life-into-shape-capturing-modeling-and-animating-3d-human-breathing}. For that level of landmark distance, the chest measurement error goes up to $6.8$cm. 

To provide more context to the results obtained by the sparse methods, we show the ambiguity error for each measurement in the following evaluation tables for a fixed maximum landmark distance of $5.64$mm, obtained as the average median landmarking manual error \cite{errors-in-landmarking-and-the-eval-of-the-acc-of-traditional-and-3d-anthropo}. As is standard practice, we also add the allowable error (AE)~\cite{AE} for manual measurements to the tables. 

% ZA MAX LM DIST OD 0.564836198464036
% ankle left circumference 0.26194711433629725
% arm length (shoulder to elbow) 0.1277391076121237
% arm right length 0.40369158011883677
% arm length (spine to wrist) 0.3169430780815996
% chest circumference 1.5523781945467774
% crotch height 0.23690564023600702
% head circumference 0.15584034893243448
% Hip circumference max height 0.14876568996282646
% hip circumference 0.7078728197977391
% neck circumference 0.16695751793973646
% height 0.25157957650501217

% ZA MAX LM DIST OD 5.64836198464036 mm
% ankle left circumference 	 2.62mm
% arm length (shoulder to elbow) 	 1.28mm
% arm right length 	 4.04mm
% arm length (spine to wrist) 	 3.17mm
% chest circumference 	 15.52mm
% crotch height 	 2.37mm
% head circumference 	 1.56mm
% Hip circumference max height 	 1.49mm
% hip circumference 	 7.08mm
% neck circumference 	 1.67mm
% height 	 2.52mm

\subsection{Metrics}

For the static datasets where the ground truth manual measurements are provided, we  use the mean absolute error (MAE) metric to evaluate our results. The MAE for a single measurement $j$ is computed as:
\begin{equation}
    MAE^j = \frac{1}{N} \sum_{i=1}^{N} \lvert gt^{j}_{i} - est^{j}_{i} \rvert,
    %y^{i}_{gt} - y^{i}_{est} \rvert
\end{equation}

\noindent where $N$ is the number of subjects, $gt^{j}_{i}$ is the ground truth measurement for subject $i$ and measurement $j$ and $est^{j}_{i}$ is the estimate of measurement $j$ for subject $i$. We further provide the average mean absolute error (aMAE)
% \begin{equation}
%     aMAE = \frac{1}{11} \sum_{j=1}^{11} MAE^{j}
% \end{equation}
which averages the MAEs over the $11$ measurements for each method.

For the dynamic and clothed test sets, where the ground truth measurements are not provided, we follow \cite{efficient-model-free-anthropometry-from-depth-data} and use the standard deviation over the difference between the measurements of each frame and the first frame.

% FOR MODEL 2024_07_11_09_42_48 trained on tsoli without bad subjects and eval on that as well
\begin{table}[t!]
\scriptsize
\begin{center}
\caption{MAE in mm on the standing and noisy test sets (male subjects). The left shows results on the standing test set, whereas the right shows results on the noisy test set. We abbreviate the measurements as: Circumferences (C.), Heights (H.) and Lengths (L.).} 
\label{tab:caesar-stand-male-res}
\begin{tabular}{|l|c|c|c|c|c|c|c|c|c|c|c|}
\cline{2-7} \cline{10-12}
      \multicolumn{1}{c|}{}     & \multicolumn{6}{c|}{ORIGINAL DATA}    &    \multicolumn{2}{c|}{} & \multicolumn{3}{c|}{NOISY DATA} \\
\cline{2-12}
      \multicolumn{1}{c|}{}     & \multicolumn{3}{c|}{dense methods}           & \multicolumn{3}{c|}{sparse methods}  &  {\tiny AE} & \multirow{ 2}{*}{\tiny Ambiguity} &  \multicolumn{3}{c|}{sparse methods} \\ 
\cline{1-7}  \cline{10-12}
Measurement              & {\tiny \cite{anthroscan}} & {\tiny \cite{scan-db}} & {\tiny \cite{tsoli-mode-based-anthropometry}} & {\tiny baseline} & {\tiny \cite{bojanic_VISAPP24}} & {\tiny Ours}  & {\tiny \cite{AE}} & & {\tiny baseline} & {\tiny \cite{bojanic_VISAPP24}} & {\tiny Ours} \\ \hline
Ankle C.     & 13.66   & 5.72    & 5.56 &  26.50   & 10.33 & 6.66               & 4     & 2.62  &25.74&10.48& 6.96  \\ 
Shoulder-elbow L. & 13.99   & 12.66   & 13.32 & 48.94   & 7.25 & 12.70          & 6   & 1.28  &49.61&7.61&  12.71   \\ 
Shoulder-wrist L. & 14.49   & 13.76   & 12.66 & 80.35   & 14.22 & 15.32         & -  & 4.04  &81.28&14.66& 16.30 \\ 
Spine-wrist L.  & 14.71   & 11.81   & 10.40 & 73.43   & 12.85 & 16.18           & -   & 3.17   &74.39&12.80&  17.13  \\ 
Chest C.   & 13.96   & 15.21   & 13.02 & 62.15   & 99.94 & 21.89                & 15    & 15.52   &64.38&100.46& 22.19 \\ 
Crotch H. & 11.01   & 9.77    & 8.36 & 13.04    & 26.73 & 8.80                  & 10     & 2.37  &12.96&31.09& 12.50 \\ 
Head C.    & 5.51    & 7.46    & 5.59 & 20.85    & 9.34 & 10.40                 & 5     & 1.56   &20.46&9.63& 10.71 \\ 
Hip C. H. & 16.50   & 18.89   & 19.05 & 27.32   & 37.80 & 20.76                 & -    & 1.49   &27.60&38.69& 21.89  \\ 
Hip C.      & 7.90    & 12.57   & 10.66 & 41.15   & 35.55 & 20.67               & 12   & 7.08   &41.99&36.64& 21.05  \\ 
Neck base C. & 21.57   & 13.33   & 13.47 & 65.26   & 14.33 & 16.99              & 11  & 1.67    &63.03&14.66& 16.81  \\
Stature               & 5.86    & 7.98    & 6.53 & 13.90    & 25.09 & 11.68     & 10 & 2.52   &14.08&25.46&  11.85   \\ \hline \hline
aMAE                     & 12.65   & 11.74   & 10.78 & 42.99   & 26.68 & 14.73  &      \multicolumn{2}{c|}{}   &43.23&27.47& 15.46 \\   \cline{1-7}  \cline{10-12}
\end{tabular}
\end{center}
\end{table}

\begin{table}[t!]
\scriptsize
\begin{center}
\caption{MAE in mm on the standing and noisy test sets (female subjects). The left shows results on the standing test set, whereas the right shows results on the noisy test set. We abbreviate the measurements as: Circumferences (C.), Heights (H.) and Lengths (L.).}
\label{tab:caesar-stand-female-res}
\begin{tabular}{|l|c|c|c|c|c|c|c|c|c|c|c|}
\cline{2-7} \cline{10-12}
  \multicolumn{1}{c|}{}     & \multicolumn{6}{c|}{ORIGINAL DATA}    &    \multicolumn{2}{c|}{} & \multicolumn{3}{c|}{NOISY DATA} \\
\cline{2-12}
      \multicolumn{1}{c|}{}     & \multicolumn{3}{c|}{dense methods}           & \multicolumn{3}{c|}{sparse methods}  &  {\tiny AE} & \multirow{ 2}{*}{\tiny Ambiguity} &  \multicolumn{3}{c|}{sparse methods} \\ 
\cline{1-7}  \cline{10-12}
Measurement              & {\tiny \cite{anthroscan}} & {\tiny \cite{scan-db}} & {\tiny \cite{tsoli-mode-based-anthropometry}} & {\tiny baseline} & {\tiny \cite{bojanic_VISAPP24}} & {\tiny Ours}  & {\tiny \cite{AE}} & & {\tiny baseline} & {\tiny \cite{bojanic_VISAPP24}} & {\tiny Ours} \\ \hline
Ankle C.             & 7.55    & 6.59     & 6.19 & 15.77   & 16.69 & 7.01                 & 4     & 2.62  & 15.50 & 16.42 & 7.04  \\ 
Shoulder-elbow L.    & 11.26   & 8.42     & 6.65 & 29.89   & 6.19 & 7.27                  & 6   & 1.28  &29.92 & 5.91 & 7.24  \\ 
Shoulder-wrist L.    & 11.67   & 10.42    & 10.05 & 57.29   & 22.24 & 9.11                & -  & 4.04  & 57.26 & 22.77 & 9.73  \\ 
Spine-wrist L.       & 13.19   & 13.40    & 11.87 & 57.04   & 12.94 & 12.70               & -   & 3.17   & 56.77 & 13.42 & 13.51  \\ 
Chest C.             & 12.43   & 13.02    & 12.73 & 33.28   & 37.67 & 20.54               & 15    & 15.52   & 33.99 & 37.03 & 20.66 \\  
Crotch H.            & 7.45    & 7.53     & 5.50 & 10.70   & 50.49 & 6.93                 & 10     & 2.37  & 10.69 & 39.79 & 14.07 \\ 
Head C.              & 7.44    & 7.45     & 5.91 & 18.59   & 9.11 & 9.70                  & 5     & 1.56   & 18.41 & 9.20 & 9.62 \\
Hip C. H.            & 17.05   & 18.96    & 18.59 & 32.70   & 36.38 & 20.00               & -    & 1.49   & 32.65 & 36.20 & 20.24 \\  
Hip C.               & 7.47    & 16.15    & 12.35 & 38.48   & 25.61 & 22.61               & 12   & 7.08   & 39.00 & 24.29 & 22.60 \\
Neck base C.         & 21.03   & 16.35    & 15.43 & 73.53   & 17.19 & 17.10               & 11  & 1.67    & 72.75 & 17.69 & 17.31 \\
Stature              & 5.60    & 10.21    & 7.51 & 13.35   & 36.85 & 9.79                 & 10 & 2.52   & 14.38 & 36.74 & 10.03 \\  \hline \hline
aMAE                 &  11.10   &  11.68    & 10.25 & 34.60   & 24.67 & 12.98    &      \multicolumn{2}{c|}{}   & 34.67 & 23.59 & 13.82 \\   \cline{1-7}  \cline{10-12}
\end{tabular}
\end{center}
\end{table}

\subsection{Standing pose} 

% compare male and female
We compare our results on the CAESAR test subjects in the standard standing A-pose. We show the results for the male and female subjects separately in Tables \ref{tab:caesar-stand-male-res} and~\ref{tab:caesar-stand-female-res}. Comparing the two tables, we can notice that the errors for the female subjects are in general lower for all the methods. The results indicate that the arm lengths for the male subjects are harder to estimate.

% comment on the dense data results
We first analyze results of methods that use dense scan data as input. Anthroscan \cite{anthroscan}, achieves the lowest MAE for the hip circumference. Anthroscan extracts the measurements directly from the dense scan data, without using a body template or relying on accurate landmark data. However, the method achieves the worst average results among methods that rely on dense data. The template-based method~\cite{scan-db} performs slightly better (on average) than Anthroscan by fitting the SCAPE \cite{scape} body model onto the scan, and using a linear model to predict the measurements from the template vertices. Tsoli et al.~\cite{tsoli-mode-based-anthropometry} improve on this by fitting a BlendSCAPE \cite{blendSCAPE} body model to the 3D scan and extracting complex features from the template to learn the measurement. By using more sophisticated features, such as a set of predefined paths, PCA coefficients of the body model fitting, and limb lengths, they achieve the best results because they considers a much greater number of local and global features.
%The best method among the approaches that require dense scan data \cite{tsoli-mode-based-anthropometry} achieves an error of $10$mm.

% comment on ours
Second, we consider works that only use sparse landmark locations as input, which avoids the requirement of dense data and template fitting. Among sparse methods, our approach achieves the best results followed by~\cite{bojanic_VISAPP24} and the baseline, respectively. The main limitation of our approach are the chest and hip circumferences which have an error over $20$ mm. However, as we see from the ambiguity column and Fig.~\ref{fig:ambiguity-lm-measurements}, these measurements have the largest ambiguity in terms of landmark locations. Note that our approach achieves results that are only $3.34$ mm less accurate on average than the best method operating on dense data.

% comment on noisy data da smo robust
When testing on the noisy test set (right of Tab.~\ref{tab:caesar-stand-male-res} and~\ref{tab:caesar-stand-female-res}), it is interesting to note that all the sparse methods, including ours, are robust to noise, and achieve only slightly worse results when landmarking errors are introduced.

\subsection{Sitting pose}
\begin{table*}[t!]
\scriptsize
\begin{center}
\caption{MAE in mm on the sitting test set. We abbreviate the measurements as: Circumferences (C.), Heights (H.) and Lengths (L.). Ambig. stands for ambiguity.}
\label{tab:caesar-sitting-res}
\begin{tabular}{|l|c|c||c|c|c||c|c||c|c|c|c|c|}
\cline{2-11}
   \multicolumn{1}{c|}{}                    & \multicolumn{5}{c||}{MALE}         & \multicolumn{5}{c|}{FEMALE}       \\ 
\cline{2-13}
   \multicolumn{1}{c|}{}                    & \multicolumn{2}{c||}{dense}         & \multicolumn{3}{c||}{sparse}  & \multicolumn{2}{c||}{dense}         & \multicolumn{3}{c|}{sparse} &   AE  & \multirow{ 2}{*}{\tiny Ambig.} \\ \cline{1-11}
Measurement & {\tiny \cite{scan-db}} & {\tiny \cite{tsoli-mode-based-anthropometry}} & {\tiny baseline} & {\tiny \cite{bojanic_VISAPP24}} & {\tiny Ours} & {\tiny \cite{scan-db}}  & {\tiny \cite{tsoli-mode-based-anthropometry}} & {\tiny baseline} & {\tiny \cite{bojanic_VISAPP24}} & {\tiny Ours} & {\tiny\cite{AE}} & \\ \hline
Ankle C.                & 5.79    & 5.22    & 29.97     & 11.81 & 7.05         & 7.00    & 6.36    & 19.03 & 31.64     & 7.57      & 4  &    2.62     \\
Shoulder-elbow L.       & 13.83   & 13.07   & 47.34     & 57.80 & 12.05         & 8.09    & 6.66    & 26.83 & 50.04     & 9.91      & 6  &    1.28    \\
Shoulder-wrist L.       & 12.75   & 12.29   & 79.14     & 122.21 & 15.21         & 10.89   & 10.49   & 54.77 & 151.11     & 10.56     & -   &   4.04    \\
Spine-wrist L.          & 11.61   & 11.26   & 67.39     & 115.07 & 18.37         & 13.16   & 12.34   & 52.45 & 164.30     & 12.72     & -   &   3.17    \\
Chest C.                & 15.40   & 13.86   & 95.77     & 49.69 & 26.63         & 13.48   & 13.05   & 84.59 & 129.33     & 21.57     & 15  &   15.52    \\
Crotch H.               & 14.19   & 14.54   & 29.85     & 149.51 & 19.87         & 11.03   & 9.66    & 16.59 & 131.48     & 17.13     & 10  &   2.37    \\
Head C.                 & 6.51    & 5.42    & 17.71     & 28.98 & 9.76         & 7.45    & 5.67    & 17.43 & 28.91     & 10.35     & 5   &   1.56    \\
Hip C. H.               & 21.41   & 20.84   & 42.27     & 179.59 & 25.17         & 23.72   & 22.26   & 29.73 & 147.54     & 27.07     & -   &   1.49    \\
Hip C.                  & 13.84   & 13.68   & 62.17     & 36.87 & 30.33         & 17.34   & 15.48   & 48.30 & 61.30     & 58.47     & 12  &   7.08    \\
Neck base C.            & 13.33   & 13.47   & 52.60     & 18.79 & 14.79         & 16.96   & 15.79   & 52.49 & 24.19     & 17.96     & 11  &    1.67   \\
Stature                 & 10.59   & 8.99    & 18.28     & 328.21 & 15.93         & 11.40   & 10.28   & 16.30 & 284.39     & 13.80     & 10  &    2.52   \\ \hline \hline
aMAE                    & 12.65   & 12.05   & 49.32     & 99.86 & 17.74         & 12.77   & 11.64   & 38.05 & 109.48     & 18.83     &   \multicolumn{2}{c}{}  \\    \cline{1-11}  
\end{tabular}
\end{center}
\end{table*}

We compare results on the sitting test set in Tab. \ref{tab:caesar-sitting-res}. The errors increase w.r.t.~the standard A-pose. The main challenge with the sitting pose is the occlusion of the back of the body due to the pose. This affects the template-based methods~\cite{scan-db,tsoli-mode-based-anthropometry} and the baseline, since the fitting relies on dense scan data. Among the sparse methods,~\cite{bojanic_VISAPP24} has only been trained on the A-pose, and is therefore not adept to estimating the measurements from posed landmarks. Our approach outperforms other sparse methods by a large margin and is only about $5mm$ less accurate on average than for the standing test set. The main reason to the degradation are the missing landmarks, as all scans in this test set suffer from missing landmarks. The landmarks are therefore transferred from the SMPL fitting for our approach and \cite{bojanic_VISAPP24}, and additionally reposed for the baseline. Compared to the best dense method, our method is only $6.44$ mm less accurate on average.

\subsection{Arbitrary pose}
In this section we analyze the pose-independence of our approach. We evaluate our method on the posed, dynamic and clothed test sets.

\subsubsection{Semi-synthetic data.}
\begin{table}[t!]
\scriptsize
\begin{center}
\caption{MAE in mm on the posed test set. We abbreviate the measurements as: Circumferences (C.), Heights (H.) and Lengths (L.).}
\label{tab:caesar-posed-res}
\begin{tabular}{|l|c|c|c||c|c|c|c|c|}
\cline{2-7}
      \multicolumn{1}{c|}{} & \multicolumn{3}{c||}{MALE}       &      \multicolumn{3}{c|}{FEMALE}      &  \multicolumn{1}{c}{}  \\ \hline
Measurement          & Baseline & \cite{bojanic_VISAPP24} & Ours  & Baseline & \cite{bojanic_VISAPP24} & Ours  & AE \cite{AE} & Ambiguity \\ \hline
Ankle C.             & 25.92    & 23.13 & 6.62  & 18.25    & 20.00 & 7.29  & 4  &    2.62 \\
Shoulder-elbow L.    & 47.45    & 76.02 & 14.86  & 35.79    & 64.07 & 8.47  & 6  &    1.28 \\
Shoulder-wrist L.    & 78.06    & 210.35 & 17.18  & 66.33    & 187.68 & 11.10  & -   &   4.04 \\
Spine-wrist L.       & 62.81    & 210.21 & 19.35  & 71.36    & 191.05 & 15.89  & -   &   3.17 \\
Chest C.             & 74.27    & 120.77 & 29.17  & 26.38    & 109.69 & 26.18  & 15  &   15.52 \\
Crotch H.            & 13.23    & 34.34 & 10.61  & 11.27    & 40.20 & 8.13  & 10  &   2.37 \\
Head C.              & 20.45    & 32.56 & 10.44  & 22.21    & 39.16 & 10.61  & 5   &   1.56 \\
Hip C. H.            & 28.28    & 72.53 & 22.99  & 32.59    & 60.61 & 20.68  & -   &   1.49 \\
Hip C.               & 41.10    & 92.59 & 20.91  & 40.66    & 62.73 & 25.39  & 12  &   7.08 \\
Neck base C.         & 63.33    & 48.21 & 17.39  & 82.48    & 41.42 & 17.08  & 11  &    1.67 \\
Stature              & 17.19    & 74.62 & 14.62  & 13.70    & 57.25 & 13.99  & 10  &    2.52 \\ \hline \hline
aMAE                 & 42.92    & 90.49 & 16.74  & 38.27    & 79.44 & 14.98  &  \multicolumn{2}{c}{} \\ \cline{1-7}
\end{tabular}
\end{center}
\end{table}

We evaluate our approach on the posed test set in Tab.~\ref{tab:caesar-posed-res}. As can be seen, both the baseline and \cite{bojanic_VISAPP24} do not have the capability to estimate the measurements from posed subjects. Contrarily, our approach achieves very similar results to the standard A-pose. This shows that our approach is robust to pose changes as long as reliable landmark input is provided (unlike in the case of the sitting pose, where landmark information was missing).

\subsubsection{Scan data.}
\begin{table}[h!]
\footnotesize
\begin{center}
\caption{Standard deviation in cm of the estimated sequence measurements for the dynamic test set.}
\label{tab:dyna-res}
\begin{tabular}{|l|c|c|c|c|}
\hline
 Measurement    & \cite{metric-regression-forests-for-human-pose-estim}+ICP & \cite{efficient-model-free-anthropometry-from-depth-data} & Ours \\  \hline
Arm L. & 1.16     & 0.77      &  0.43    \\
Hip C. & 1.78     & 1.70      &   0.65    \\ \hline 
\end{tabular}
\end{center}
\end{table}
\begin{table}[h]
\begin{center}
\scriptsize
\caption{The standard deviation in mm on the clothed test set.}
\label{tab:kinovis-res}
\begin{tabular}{|l|c|l|c|}
\hline
Measurement    & Std (mm) & Measurements & Std (mm) \\ \hline
% Ankle C.             &        1.03                 & Head C.      &           3.51              \\
% Shoulder to elbow L. &         0.79                & Hip C. H.    &             4.19            \\
% Shoulder to wrist L. &          2.30               & Hip C.       &            3.93             \\
% Spine to wrist L.    &           1.42              & Neck base C. &             2.53            \\
% Chest C.             &           4.33              & Stature      &             9.57            \\ 
% Crotch H.            &            2.71             &              &                         \\ \hline
Ankle C.             &        2.05                 & Head C.      &           4.19              \\
Shoulder to elbow L. &         2.58                & Hip C. H.    &             8.45            \\
Shoulder to wrist L. &          3.34               & Hip C.       &            7.51             \\
Spine to wrist L.    &           4.34              & Neck base C. &             3.90            \\
Chest C.             &           12.51              & Stature      &             11.29            \\ 
Crotch H.            &            8.11             &              &                         \\ \hline
\end{tabular}
\end{center}
\end{table}

% STD per measurement in cm for 5640
% ankle left circumference          0.205987
% arm length (shoulder to elbow)    0.258655
% arm right length                  0.334288
% arm length (spine to wrist)       0.434812
% chest circumference               1.251222
% crotch height                     0.811622
% head circumference                0.419736
% Hip circumference max height      0.845799
% hip circumference                 0.751745
% neck circumference                0.390675
% height                            1.129707

We evaluate our approach on the dynamic test set in Tab.~\ref{tab:dyna-res} and the clothed test set in Tab.~\ref{tab:kinovis-res}. We compute the standard deviation of the difference between each frame and the first frame. This way we can analyze the consistency of our approach w.r.t.~a dynamic change in pose, or even change in pose obstructed by clothing. As we can see from both tables, our approach behaves consistently with slight variations to the measurements during the dynamic movement. This indicates that even though our approach has an error of around $15$ mm for a subject on average (seen in previous results), it consistently predicts the same measurements for a single subject, achieving pose-independence. For the dynamic test set, Tab.~\ref{tab:dyna-res} further shows that our approach slightly outperforms the state-of-the-art dynamic method on the two measurements for which results are available.

\section{Conclusion}
The current pipeline for 3D anthropometry consists of scanning the subject in the static A-pose without any movement. However, the A-pose is not easy to maintain during a scanning process that can last up to a couple of minutes. This constraint affects current state-of-the-art template-based methods that rely on dense geometry data. Additionally, this constraint makes it impossible to develop a digital anthropometry method for subjects unable to assume the A-pose, such as those with injuries or disabilities. We propose a robust approach that estimates body measurements from any given pose using only sparse landmark data. We thoroughly test our method and show that it achieves comparable results to the state-of-the-art methods operating on dense scan data in the standard A-pose, but has the capability of estimating the body measurements from landmark data only, acquired in any pose. We address the lack of open-source 3D anthropometry methods by making our method available to the research community.

\newpage
\noindent \textbf{Acknowledgement} \\
This work has been supported by the Croatian Science Foundation under the projects IP-2018-01-8118 and DOK-2020-01, as well as the European Union - NextGenerationEU. Views and opinions expressed are however those of the author(s) only and do not necessarily reflect those of the European Union or European Commission. Neither the European Union nor the European Commission can be held responsible for them. We thank Antoine Dumoulin and Rim Rekik Dit Nkhili for helping with the SMPL fitting and Briac Toussaint for helping with the visualizations. In memory of our dear colleague and friend, Tomislav Pribanić.

\bibliographystyle{splncs04}
\bibliography{main}
\end{document}